\documentclass[10pt,twocolumn,letterpaper]{article}

\usepackage{cvpr}              

\usepackage{graphicx}
\usepackage{amsmath}
\usepackage{amssymb}
\usepackage{booktabs}
\usepackage{enumitem}
\usepackage{multirow}
\usepackage{xcolor}
\usepackage{colortbl}
\usepackage{comment}
\usepackage[pagebackref,breaklinks,colorlinks]{hyperref}
\newcommand{\cellc}{\cellcolor{white!15}}

\usepackage[capitalize]{cleveref}
\crefname{section}{Sec.}{Secs.}
\Crefname{section}{Section}{Sections}
\Crefname{table}{Table}{Tables}
\crefname{table}{Tab.}{Tabs.}

\def\cvprPaperID{1292} 
\def\confName{CVPR}
\def\confYear{2023}

\begin{document}

\title{Enhanced Training of Query-Based Object Detection via \\ Selective Query Recollection}

\author{Fangyi Chen$^1$ \quad Han Zhang$^1$ \quad Kai Hu$^1$ \quad Yu-Kai Huang$^1$ \quad Chenchen Zhu$^2$ \quad Marios Savvides$^1$\\
Carnegie Mellon University$^1$ \qquad Meta AI$^2$\\
{\tt\small \{fangyic,hanz3,kaihu,yukaih2,marioss\}@andrew.cmu.edu \quad chenchenz@fb.com }
}
\maketitle

\begin{abstract}
    This paper investigates a phenomenon where query-based object detectors mispredict at the last decoding stage while predicting correctly at an intermediate stage. We review the training process and attribute the overlooked phenomenon to two limitations: 
    lack of training emphasis and cascading errors from decoding sequence.
    We design and present Selective Query Recollection (SQR), a simple and effective training strategy for query-based object detectors. 
    It cumulatively collects intermediate queries as decoding stages go deeper and selectively forwards the queries to the downstream stages aside from the sequential structure. 
    Such-wise, SQR places training emphasis on later stages and allows later stages to work with intermediate queries from earlier stages directly. SQR can be easily plugged into various query-based object detectors and significantly enhances their performance while leaving the inference pipeline unchanged. As a result, we apply SQR on Adamixer, DAB-DETR, and Deformable-DETR across various settings (backbone, number of queries, schedule) and consistently brings $1.4\sim2.8$ AP improvement. Code is available at \url{https://github.com/Fangyi-Chen/SQR}
\end{abstract}

\section{Introduction}
\label{sec:intro}

\begin{figure}
    \centering
    \includegraphics[width=1.0\columnwidth]{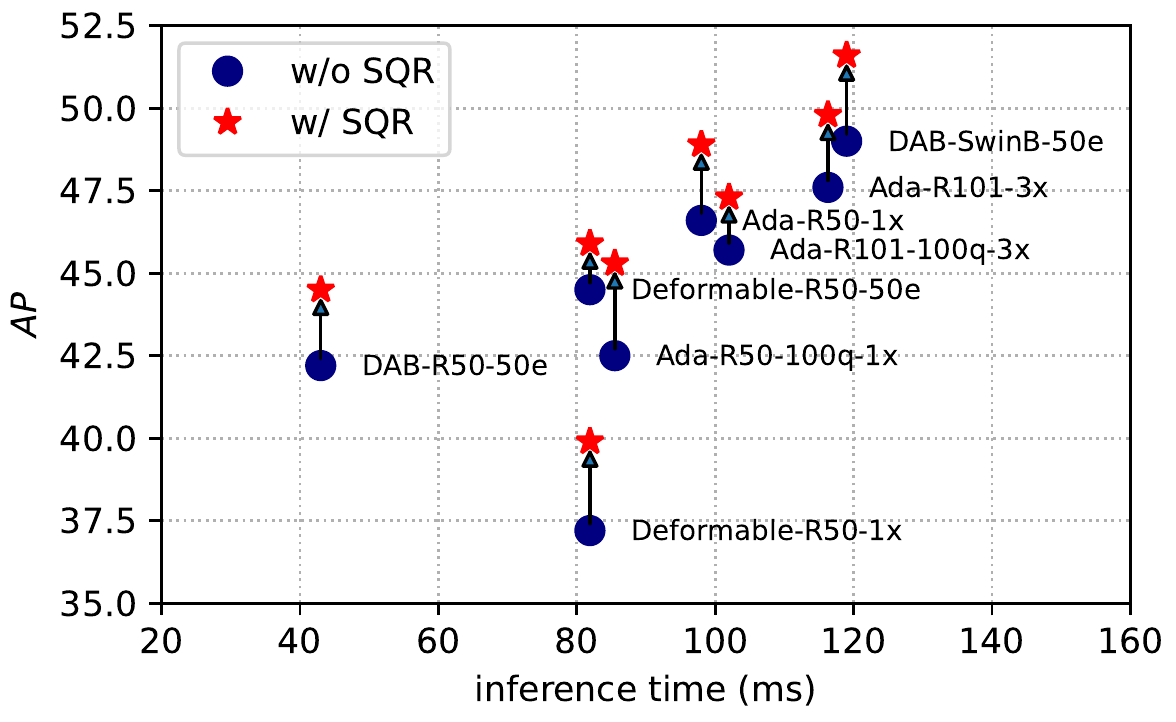}
    \caption{The inference speed and AP for various networks on the MS-COCO val set. The red stars are the results trained with SQR. The blue circles are the results of baselines without SQR. SQR enhances the training of query-based object detectors while leaving the inference pipeline unchanged.}
    \label{fig:speedvsap}
\end{figure}

Object detection is a long-established topic in computer vision aiming to localize and categorize objects of interest. Previous methods \cite{Girshick2014RichFH, Girshick2015FastR, Ren2015FasterRT, Liu2016SSDSS, Redmon2016YouOL, haibao1, haibao2, Lin2017FocalLF, Zhu2020SoftAO, Chen2020NCMSTA, Zhang2020BridgingTG, Tian2019FCOSFC, Kong2019FoveaBoxBA, Zhang2021VarifocalNetAI, feng2021tood} rely on dense priors tiled at feature grids so as to detect in a sliding-window paradigm, and have dominated object detection for the recent decade, but these methods fail to shake off many hand-crafted processing steps such as anchor generation or non-maximum suppression, which block end-to-end optimization.

Recent research attention has been geared towards query-based object detection \cite{Carion2020EndtoEndOD, Liu2022DABDETRDA, Meng2021ConditionalDF, Wang2022AnchorDQ, Zhu2021DeformableDD, Li2022DNDETRAD, Sun2021SparseRE} since the thriving of transformer \cite{Vaswani2017AttentionIA} and DETR \cite{Carion2020EndtoEndOD}. By viewing detection as a direct set prediction problem, the new archetype represents the set of objects using a set of learnable embeddings, termed as queries, which are fed to a decoder consisting of a stack (typically six) of decoding stages. Each stage performs similar operations: (1) interacting queries with image features via an attention-like mechanism, so the queries are aggregated with valuable information that represents objects; (2) reasoning the relation among all queries so that global dependency on objects co-occurrence and duplicates could be captured; (3) interpreting bounding box and category from each query by a feed forward network. Queries are sequentially processed stage-by-stage, and each stage is formulated to learn a residual function with reference to the former stage's output, aiming to refine queries in a cascaded style.

\begin{figure*}[t]
    \centering
    \includegraphics[width=1.0\textwidth]{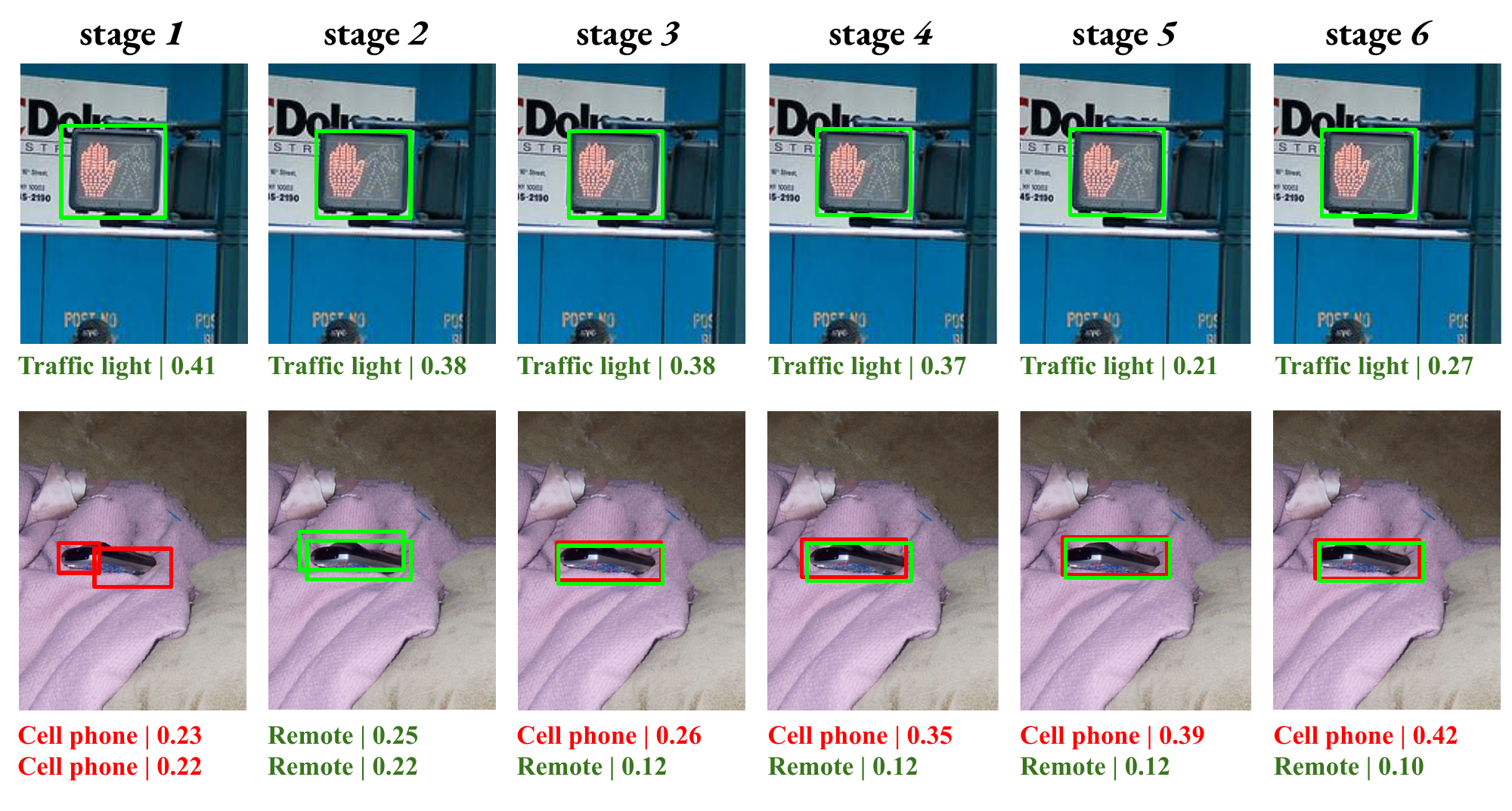}
    \caption{Are query-based object detectors always enhancing predictions stage-by-stage? The \emph{traffic light} at stage 1 gets a confident score of 0.41, while from stage 2 to 5 the confidence gradually decreases to 0.21 (Upper); the \emph{remote} at stage 3 was wrongly classified as a cell phone, and from stage 3 to 6 the mistake was amplified from 0.26 to 0.42 (Lower). The visualization is acquired from Adamixer-R50 (42.5 AP) tested on COCO val set.}
    \label{fig:fig1}
\end{figure*}

As such wise, the decoding procedure implies that detection should be stage-by-stage enhanced in terms of IoU and confidence score. Indeed, monotonically improved AP is empirically achieved by this procedure. However, when visualizing the stage-wise predictions, we surprisingly observe that decoder makes mistakes in a decent proportion of cases where the later stages degrade true-positives and upgrade false-positives from the former stages. As shown in Fig.\ref{fig:fig1}, the traffic light at stage 1 gets categorical confidence of 0.41, while from stage 2 to 5 the confidence gradually decreases to 0.21; the remote at stage 3 was wrongly classified as a cell phone, while from stage 3 to 6 the error was exacerbated from 0.26 to 0.42. We present a more detailed statistic in Section \ref{sec:motivation}.


This phenomenon inspires us to review the current training strategy and bring two conjectures. 
\textbf{Firstly}, the responsibility that each stage takes is unbalanced, while supervision applied to them is analogous. An early stage could make mistakes without causing too much impact because it gets chances to be corrected later, and the later stages are more responsible for the final prediction. But during training, all of these stages are supervised in an equivalent manner and there lacks such a mechanism that places particular training emphasis on later stages.  
\textbf{Secondly}, due to the sequential structure of the decoder, an intermediate query refined by a stage - no matter whether this refinement brings positive or negative effects - will be cascaded to the following stages, while the query prior to the refinement never gets an opportunity to be propagated forward even though it emerges unscathed and might be more representative than the refined one. The cascading errors increase the difficulty of convergence and the sequential structure impedes the later stages from seeing prior queries during training.



Based on these intuitions, we present Query Recollection (QR) as a training strategy for query-based object detectors. It cumulatively collects intermediate queries as stages go deeper, and feeds the collected queries to the downstream stages aside from the sequential structure. By each stage, the new add-ins alongside the original inputs are independently treated among each other, so the attentions and losses are calculated individually. In such a manner, QR enjoys two key features: \textbf{(1)} The number of supervision signals per stage grows in geometric progression, so that later stages get more supervision than the former ones, for example, the sixth stage got 32 times more supervision than the first; \textbf{(2)} Later stages get chance to view the outputs beyond its neighboring stage for training, which mitigates the potential impact due to cascading errors. We further discover that \textit{selectively} forward queries to each stage, not with the entire query collection but only those from the prior two stages, can raise the number of supervision in a Fibonacci sequence which halves the extra computing cost and brings even better results. We name it Selective Query Recollection (SQR).

Our contributions are summarized in three folds: \textbf{(1)} We quantitatively investigate the phenomenon where query-based object detectors mispredict at the last decoding stage while predicting correctly at an intermediate one. \textbf{(2)} We attribute the overlooked phenomenon to two training limitations, and propose a simple and effective training strategy SQR that elegantly fits query-based object detectors. \textbf{(3)} We conduct experiments on Adamixer, DAB DETR, and Deformable DETR across various training settings that verify its effectiveness (Fig.\ref{fig:speedvsap}).

\section{Related Work}
\label{sec:relatedwork}
\subsection{Training Strategy for Object Detection}
%
%

Detectors based on dense priors have been dominating the community for decades. The abstract concept anchor box or anchor point \cite{Ren2015FasterRT, Tian2019FCOSFC} aims to match with ground truth (GT) objects depending on their Intersection-over-Union (IoU) values or other advanced soft scoring factors \cite{Lin2017FocalLF, Tian2019FCOSFC, Kim2020ProbabilisticAA, Zhu2020SoftAO, feng2021tood}. Among anchor-based detectors, multi-stage models iteratively refine bounding box and category stage-by-stage. A typical example is Cascade RCNN \cite{Cai2018CascadeRD} which is based on the design that the output of an intermediate stage is sampled and re-labeled to train the next stage with increasing IoU thresholds, so these stages are guaranteed to be progressively refined. Recently, DETR \cite{Carion2020EndtoEndOD} starts a family of end-to-end query-based models where object detection is regarded as a set prediction problem. 
To train DETR, a predefined number of object queries are matched to either a ground-truth or background by solving the Hungarian Matching problem. The queries are refined by several decoder stages similar to Cascade RCNN, and each intermediate stage is supervised by the matching results.  

\subsection{Query-Based Object Detection}
Recently, many algorithms have been following the idea of DETR. Deformable DETR \cite{Zhu2021DeformableDD} proposes a deformable attention module that alleviates the aforementioned issues and massively improves the convergence speed by a factor of 10. 
Conditional DETR \cite{Meng2021ConditionalDF} decouples the object query into content query and spatial query in the decoder cross-attention module and learns a conditional spatial query from the decoder embedding to enable fast learning of the distinctive extremity of the ground-truth objects. 
Anchor-DETR \cite{Wang2022AnchorDQ} formulates the object queries as anchor points such that each object query may only focus on a certain region near its anchor point. Many future works are inspired by this design.
DAB-DETR \cite{Liu2022DABDETRDA} dives deep into the role of object queries. It directly uses anchor box coordinates as spatial query to speed up training. The model benefits from the spatial prior by modulating the positional attention map using the width and height of the box. 
DN-DETR \cite{Li2022DNDETRAD} further improves the convergence speed and query matching stability of DAB-DETR with the help of the Ground Truth denoising task.  
Adamixer \cite{Gao2022AdaMixerAF} re-designs the query-key pooling mechanism by letting the query adaptively attend to the meaningful regions of the encoded features directly through bilinear sampling. An MLP-Mixer then dynamically decodes the pooled features into final predictions. 
DETA (Detection Transformers with Assignment)\cite{NMSback} proposes a novel overlap-based assignment that enables one-to-many assignments to DETR. 

Despite their significant improvement on DETR, they focus little on the training emphasis and cascading errors. We propose SQR to pay attention to the two problems.


\section{Motivation}
\label{sec:motivation}
Are query-based object detectors always predicting the optimal detections at the last stage? Table.\ref{tab:stagewise_map} shows that AP gradually increases with stages going deeper, indicating improved predictions on a general scale. While the observation in Fig.\ref{fig:fig1} implies that simple AP results are insufficient for an in-depth analysis of that question.

\begin{table}[]
    \centering
    \begin{tabular}{c|c|c|c|c|c|c}
        \toprule[1pt]
        Model & S1 & S2 & S3 & S4 & S5 & S6 \\ \midrule
        Deformable & 38.4 & 42.2 & 43.7 & 44.2 &44.4 & 44.5 \\ 
        Adamixer & 15.1  & 30.3 &37.7 &40.6&42.1&42.5 \\ 
        \bottomrule
    \end{tabular}
    \caption{Stage-wise AP results. Deformable DETR(300 queries) and Adamixer(100 queries) are trained using official implementation, both of their decoders consist of 6 stages, and per stage AP is reported on COCO val set.}
    \label{tab:stagewise_map}
\end{table}

\begin{figure}
    \centering
    \includegraphics[width=\columnwidth]{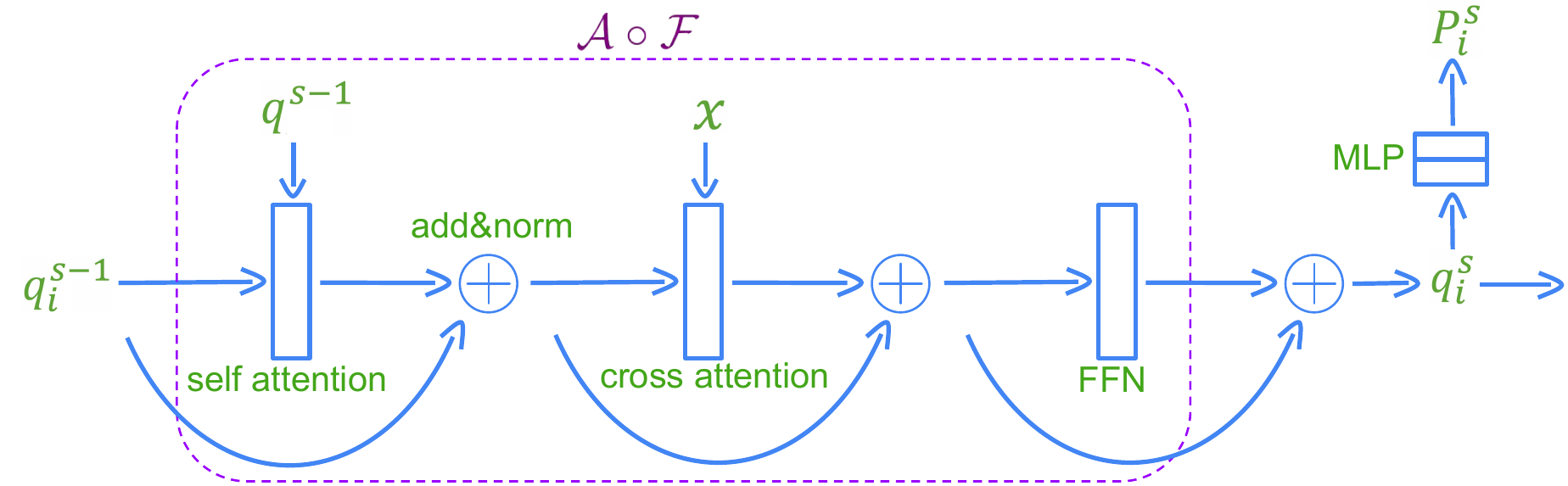}
    \caption{Typical structure of a single decoding stage.}
    \label{fig:preliminary}
\end{figure}

\textbf{Preliminary.} Queries are updated successively. A typical structure of a decoding stage is illustrated in Fig.\ref{fig:preliminary}. An initial query $q_i^0$ ($i\in N$ $=\{1,2,...,n\}$) is an embedding trained through back-propagation, and $n$ is the total number of initial queries. During inference, the first stage updates the initial query by adding a residual term to it, producing intermediate query $q_i^1$, followed by later stages sequentially updating the intermediate query in the same way. The procedure can be formulated as 
\begin{equation}
\label{eq:1}
    q_i^{s} = D^{s}(q_i^{s-1}, q^{s-1}, x)
    = q_i^{s-1} + (\mathcal{A}\circ\mathcal{F})(q_i^{s-1}, q^{s-1}, x)
\end{equation}

\noindent $D^s$ is a decoding stage where $s$ is stage index; $q_i^s$ is the $i_{th}$ query at stage $s$; $q^s$ is a set of queries $q^s=\{q_i^s|i\in N\}$; $(\mathcal{A}\circ\mathcal{F})$ stands for a bundle of modules including self and cross attention and feed forward network; $x$ means features; and LayerNorm \cite{Ba2016LayerN} that applied on each module is omitted for simplicity. Afterward, $q_i^s$ predicts an object $P_i^s$ via two multi-layer perceptrons for classification and regression: 

\begin{equation}
\label{eq:2}
    P_i^s = (~MLP_{cls}(q_i^s), ~~ MLP_{reg}(q_i^s)~)
\end{equation}

\noindent $P_i^{1\sim 6}$ are predicted by the $q_i^{1\sim 6}$ rooted in $q_i^0$, where $P_i^6$ is the expected outcome and $P_i^{1\sim 5}$ are intermediate results.
$P_i^s$ is regarded as a true-positive towards a ground-truth $G$ only if the IoU($P_i^s$, G) exceeds a threshold, its category matches with $G$, and the categorical score is ranked as the highest among all other counterparts.

\begin{table}[]
    \centering
    \begin{tabular}{c|c|c|c}
        \toprule[1pt]
        Model       & TP Threshold & TP F Rate & FP E Rate \\ 
        \midrule
        Deformable & IoU$>$0.50 & 51.4$\%$ & 55.7$\%$ \\ 
        DETR & IoU$>$0.75 & 49.5$\%$ & 55.9$\%$ \\ \midrule
        \multirow{2}{*}{Adamixer} & IoU$>$0.50 & 28.6$\%$ & 50.8$\%$ \\
         & IoU$>$0.75 & 26.7$\%$ & 51.2$\%$  \\
        \bottomrule
    \end{tabular}
    \caption{True-positive Fading Rate and False-positive Exacerbation Rate. }
    \label{tab:fade_grace}
\end{table}

\textbf{Investigation.} We study the stage-wise testing results by first defining two rates: (1) If $P_i^6$ is a true-positive (TP) towards a ground-truth $G$, we check whether $P_i^{1\sim 5}$ generate a better TP \textit{\textbf{towards the same G that has higher IoU $\&$ higher category score}} than $P_i^6$. The occurrence rate is denoted as \textit{TP fading rate}. (2) If $P_i^6$ is a false-positive (FP), we check whether $P_i^{1\sim 5}$ produce a FP \textit{\textbf{but with lower category score}} than $P_i^6$. The occurrence rate is denoted as \textit{FP exacerbation rate}. 

The statistics are shown in Table.\ref{tab:fade_grace}. We can see that TP fading rate achieves $50\%$ and $27\%$ on the two models. Considering the strict constraints it holds, the TP fading rate reaches an impressively high level. Deformable DETR is much higher than Adamixer due to the smaller AP gap among stages, thus, in many cases, earlier stages are more likely to outperform the last stage. FP exacerbation holds looser constraints and is easier to be satisfied, its rate is above $50\%$ on the two models. Higher TP threshold implies higher quality of $P_i^6$ and it is harder to find qualifiers in $P_i^{1\sim 5}$, but the two rates are similar with 0.75 as with 0.5, pointing out the consistency of this phenomenon. We also observe that more than half of the occurred cases are \textit{marginally triggered}, i.e. the predictions from the triggered $P_i^{1\sim 5}$ are only marginally better than the sixth. This is a further reason why the deformable DETR has those high rates - the results from the 5th and 6th stages are extremely visually close. In dominated number of cases, the final stage is (one of) the best, after all, its mAP is the highest. 

When the conditions of the two rates establish, we further replace $P_i^6$ with the optimal prediction found in $P_i^{1\sim 6}$ and measure the AP. On Deformable DETR, AP grows from \textbf{44.5 AP} to \textbf{51.7 (+7.2 AP)}; on Adamixer, AP grows from \textbf{42.5 AP} to \textbf{53.3 (+10.7 AP)}, demonstrating huge potential in query-based detectors yet to be mined.  

\textbf{Conclusion}. This reveals that models frequently predict the optimum at intermediate stages instead of the last one. We view the problem from the training's perspective and identify \textit{the lack of training emphasis} and \textit{the cascading errors from query sequence} as two obstacles impeding the occurrence of the most reliable predictions in the last stage, elaborated in Section \ref{sec:intro}.

\section{Query Recollection}
\label{sec:QR}

\subsection{Expectancy}
We desire such a training strategy that embraces:
\setlist[itemize]{noitemsep, nolistsep}
\begin{itemize}
    \item Uneven supervision applied to decoding stages that places emphasis on later ones, enhancing later stages for better final outcomes.
    \item A variety of early-stage queries directly introduced to later stages, mitigating the impact of cascading errors.
\end{itemize}


To this end, we design a concise training strategy coined as Query Recollection (QR). Compared with prior arts, it collects intermediate queries at every stage and forwards them along the original pathway. Dense Query Recollection (DQR) is the fundamental form and Selective Query Recollection (SQR) is an advanced variant. 

\begin{figure}
    \centering
    \includegraphics[width=1.05\columnwidth]{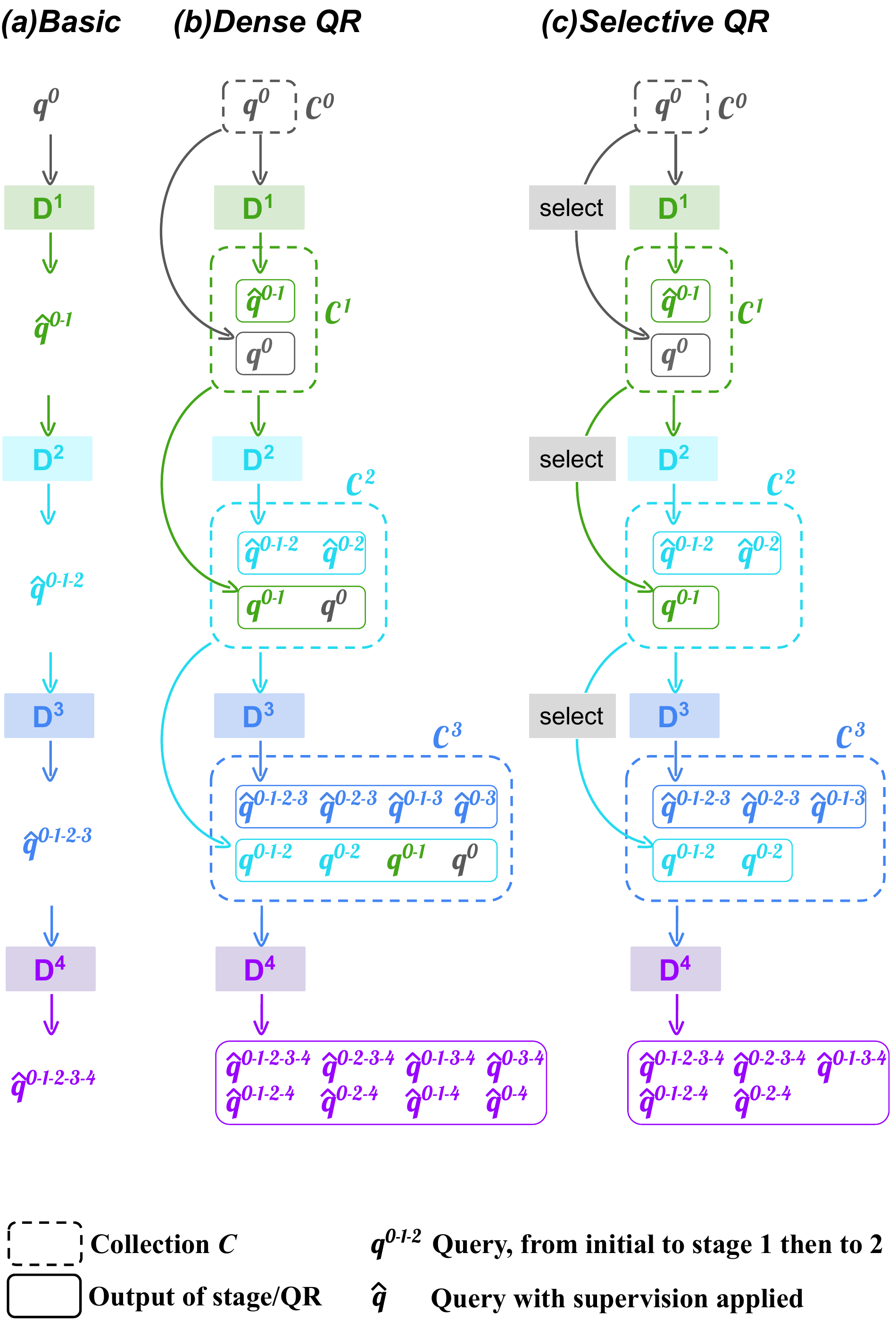}
    \caption{(a). Basic process for decoding queries stage by stage, applied in both training and testing. (b). Dense query recollection. (c). Selective query recollection.}
        \label{fig:qr}
\end{figure}

\subsection{Dense Query Recollection}
\textbf{Notation.} The process of single decoding stage (self-attention, cross-attention, FFN), the ground-truth assignment, and loss calculations are applied within a set of queries $\{q_{i}|i\in \{1,2,...,n\}\}$, where n is typically 100/300/500. We regard the set of queries as a basic unit in our method and generally denote as $q$.

\textbf{Basic Pathway.} Query along the basic pathway is refined by all stages. We illustrate it in Fig.\ref{fig:qr}(a). Taking a 4-stages decoder as an example, we denote $q^{0\text{-}1\text{-}2\text{-}3\text{-}4}$ as the final query that is refined by all stages.  So the basic $\mathcal{PT}$ finally produces

\begin{eqnarray}
\label{eq:denseQR0}
    q^{0\text{-}1\text{-}2\text{-}3\text{-}4} & = & D^{4}(D^{3}(D^{2}(D^{1}(q^0)))) \\
    &  = & \mathcal{PT}^{1\text{-}2\text{-}3\text{-}4}(q^0)
\end{eqnarray}

During training, the queries from each stage, i.e. $q^{0\text{-}1}$, $q^{0\text{-}1\text{-}2}$, $q^{0\text{-}1\text{-}2\text{-}3}$, and $q^{0\text{-}1\text{-}2\text{-}3\text{-}4}$ are independently followed by Hungarian Assignment that matches ground-truth with $q$ in a one-to-one manner, and then followed by loss calculation for supervision. In Fig.\ref{fig:qr}, we mark those $q$ that require supervision as $\hat{q}$. Along the basic pathway, the number of $\hat{q}$ at each stage is 1.

\textbf{DQR Formulation.} We densely collect every intermediate $q$ and independently forward them to every downstream stage, as illustrated in Fig.\ref{fig:qr}(b). After each stage, a collection $C$ is formed where the number of $q$ grows geometrically, namely $2^{s}$ at $s_{th}$ stage. Formally,

\begin{equation}
\label{eq:denseQR1}
    C^0 = \{q^0\} 
\end{equation}
\begin{equation}
\label{eq:denseQR2}
    C^s = \{D^s(q)|q\in C^{s-1}\} \cup C^{s-1}
\end{equation}

In a collection $C$, half of queries inside are newly generated by the current stage, i.e. from $\{D^s(q)|q\in C^{s-1}\}$, while another half are from previous stages, $C^{s-1}$. Separately, for each $q$ in the former half, we apply Hungarian assignment and loss calculation, so the number of supervision signals grows in geometric progression as well, namely $2^{s-1}$ at $s_{th}$ stage.

Such-wise, Dense Query Recollection satisfies our expectancy where the number of supervision signal for each stage grows as (1,2,4,8,16,32), meanwhile, all prior queries would be visible in all later stages.

During inference, we only use the basic pathway, so the inference process is untouched. For a standard 6-stage decoder, the pathway is $\mathcal{PT}^{1\text{-}2\text{-}3\text{-}4\text{-}5\text{-}6}$

\subsection{Selective Query Recollection}
The last proposal empirically enhances training, but the query collection process is indiscriminate, which brings two defects:
First, the geometrical growth of the number of $\hat{q}$ and their attention/loss calculations cost a lot. Second, if we input an early $q$ that skips too many stages to a far-away late stage, the potential benefit could be overshadowed by the huge learning gap between the stages and query. For instance, if the initial $q^0$ is directly introduced to stage 6 and produces $\hat{q}^{0\text{-}6}$, this query would have the highest difficulty among all queries at stage 6 and the calculated loss would dominate the overall losses. So we are inclined to selectively collect intermediate $q$ rather than densely collect all of them. 

\textbf{Selection.} To find a better query recollection scheme, we further conduct a detailed analysis on the \textit{TP Fading Rate} and \textit{FP Exacerbation Rate} introduced in Section \ref{sec:motivation}. 

\begin{table}[]
\centering
    \begin{tabular}{c|c|c|c|c|c}
        \toprule[1pt]
        stage          & 1   & 2 & 3 & 4 & 5   \\ 
        \midrule
        TP F Rate($\%$)&  1.2  & 4.4 & 8.5 & 12.4 &16.9   \\ 
        FP E Rate($\%$)&  14.3 & 18.6 & 20.8 & 24.5 & 30.2  \\
        \midrule[1pt]
        stage    &  1$\sim$3   & 4$\&$5 & 3$\sim$5 & 2$\sim$5 & 1$\sim$5  \\ 
        \midrule
        TP F Rate($\%$)&   11.2 & 23.9 & 26.9 & 28.3 & 28.6 \\ 
        FP E Rate($\%$)&   32.4 & 40.8 & 45.3 & 48.5 & 50.8 \\
        \bottomrule
    \end{tabular}
    \caption{Stage-wise TP Fading Rate and FP Exacerbation Rate with Adamixer.}
    \label{tab:dif_s4rates}
\end{table}

Recall the spirits of the two rates where we seek from $P_i^{1\sim 5}$ for an alternative that is better than $P_i^6$ , we want to investigate which specific intermediate stage/stages contribute the most. 
Concretely, if $P_i^6$ is a true-positive towards a ground-truth $G$, we separately check each stage whether generating a better TP; similarly, if $P_i^6$ is a false-positive, we separately check each stage whether generating a better FP. The results are summarized in Table \ref{tab:dif_s4rates}. We find that the majority of alternatives of $P_i^6$ are from stage 4$\&$5, where the corresponding TP fading rate and FP exacerbation rate reach 23.9\% and 40.8\%, respectively, which are close to the results of stages 1 $\sim$ 5. While stage 1 $\sim$ 3 together only produces 11.2\% and 32.4\%. 

The above analysis implies that the queries from \emph{the adjacent stage} and \emph{the stage prior to the adjacent stage} are more likely to bring positive effects. We intuitively follow the observations and selectively operate collection along the basic pathway: before starting stage $D^s$, we collect $q$ from the 2 nearest stages, i.e. $D^{s-1}$ and $D^{s-2}$ as the input of $D^s$. 

\textbf{SQR Formulation.} The Selective Query Recollection can be formulated as
\begin{eqnarray}
\label{eq:denseSQR1}
    C^0 = \{q^0\} & & C^1 = \{q^0, q^{0\text{-}1}\} 
\end{eqnarray}
\begin{eqnarray}
\label{eq:denseSQR3}
C^s & = & \{D^s(q)|q\in C^{s-1}\} \cup select(C^{s-1})\\
 & = & \{D^s(q)|q\in C^{s-1}\} \cup \{D^{s-1}(q)|q\in C^{s-2}\}
\end{eqnarray}

Such wise, Selective Query Recollection (Fig.\ref{fig:qr}(c)) still satisfies our expectancy, and the number of supervision signals grows in a Fibonacci sequence (1,2,3,5,8,13). Compared to the Dense Recollection, to a great extent, SQR reduces the computing burden and we observe that SQR even outperforms the dense counterpart in terms of precision. This verifies our assumption that a $q$ skipping too many stages might be noise for remote stages overshadowing its positive effects. 

\textbf{Recollection Starting Stage.} Above we collect queries starting from stage 1. Instead, we can practically vary this starting stage, and this will further reduce the total queries in each collection and further reduce the computing burden. E.g., if starts SQR from stage 2, the Fibonacci sequence will start from stage 2 and result in (1,1,2,3,5,8); if starts from stage 3, result in (1,1,1,2,3,5). The starting stage is regarded as a hyper-parameter for SQR.

\section{Experiments}
\label{sec:exp}

We conduct our experiments on the MS-COCO \cite{Lin2014MicrosoftCC} detection track using the MMDetection \cite{mmdetection} and Detrex \cite{ren2022detrex} code-bases. All models are trained on \textbf{train2017} split. Unless otherwise specified, models are trained and tested with image scale of 800 pixels, where AdamW optimizer with a standard 1x schedule (12 epochs) is utilized for training. For ablation study and analysis, Adamixer \cite{Gao2022AdaMixerAF} with R50 \cite{He2016DeepRL} is chosen because of its good performance and fast convergence speed (42.5 AP with 1x schedule). We further apply SQR on other detectors to verify its effectiveness. 

\subsection{Ablation Study}
\textbf{Baseline vs DQR vs SQR.} Table.\ref{tab:baseline_vs_qr} shows that both DQR and SQR improve the baseline by a large margin. DQR reaches 44.2 (+1.7 AP) while SQR reaches a slightly high result 44.4 (+1.9 AP). Note that SQR is much more efficient than DQR. As shown in Table.\ref{tab:startingstage}, under the same training setting, SQR cuts down a great amount of training time of DQR and still achieves equal or higher AP. 

\begin{table}[]
    \centering
    \begin{tabular}{c|c|c|c|c|c|c}
    \toprule[1pt]
        Methods & AP   & AP\textsubscript{50}    & AP\textsubscript{75}  & AP\textsubscript{S}   & AP\textsubscript{M}   & AP\textsubscript{L}  \\ \midrule
        
    Baseline       & 42.5 & 61.5 & 45.6 & 24.6  & 45.1  & 59.2 \\
      DQR          & 44.2 & 62.8 & 47.9 & 26.7  & 46.9  & 60.5 \\
      SQR          & 44.4 & 63.2 & 47.8 & 25.7  & 47.4  & 60.2 \\
       \bottomrule
    \end{tabular}
    \caption{AP comparison among Baseline, DQR, and SQR}
    \label{tab:baseline_vs_qr}
\end{table}

\textbf{Varying Starting Stage of SQR.} We present how SQR performs when varying the starting stage in Table.\ref{tab:startingstage}. The best performance is acquired when query recollection starts at stage 1 but with the most computational cost. We can see that starting at stage 2 performs similarly to starting at stage 1 while the computing burden is decently reduced. With recollection starting later, the benefits from SQR decrease as expected since the recollected queries from early stages become fewer and training emphasis gets gradually balanced.



\begin{table}[]
    \centering
    \begin{tabular}{c|c|c|c|c}
    \toprule[1pt]
    Method & Start Stage   & Train Time &  AP   & AP\textsubscript{50}     \\ \midrule
    Baseline &  -    &1x(5hours)&  42.5   &  61.5      \\
    Baseline &  -    &  2x   &  42.5   &  61.3      \\
    Baseline &  -    &  3x   &  42.5   &  61.4      \\
    DQR      &  -    &  2.24x&  44.2   &  62.8      \\
    SQR      &  1    &  1.57x&  44.4   &  63.2      \\
    SQR      &  2    &  1.34x&  44.2   &  63.0      \\
    SQR      &  3    &  1.18x&  43.8   &  62.3      \\
    SQR      &  4    &  1.07x&  42.9   &  61.4      \\
    \bottomrule
    \end{tabular}
    \caption{Further comparison among Baseline, DQR, and SQR with different starting stage in terms of training time and AP.}
    \label{tab:startingstage}
\end{table}

\textbf{TP Fading Rate And FP Exacerbation Rate.}
We present Table.\ref{tab:ablation_FErates} to verify that TP fading rate and FP exacerbation rate decrease due to the training effect when applied SQR. Specifically, TP fading rate decreases from 28.6\% to 23.3\% and from 26.7\% to 21.1\% across two IoU thresholds. FP exacerbation rate downgrades from 50.8\% to 47.3 and 51.2\% to 47.0\%.   

\textbf{Another Treatment on Collected Queries.} DQR or SQR treats each set of queries $q$ independently as $q$ undergoes its own attention and Hungarian matching without explicitly interacting with each other. We try an alternative: at each stage, the collected queries are all concatenated together, so each stage generates only one set of $q$ but the number of $q_i$ grows as DQR or SQR. We found this brings +0.3 AP increase compared to the baseline.

\begin{table}[]
    \centering
    \begin{tabular}{c|c|c|c}
        \toprule[1pt]
        Method       & TP Threshold & TP F Rate & FP E Rate \\ 
        \midrule
        Baseline & IoU$>$0.50 & 28.6$\%$   & 50.8$\%$ \\
        SQR      & IoU$>$0.50 &  23.3 $\%$ & 47.3 $\%$ \\
        \midrule
        Baseline & IoU$>$0.75 & 26.7$\%$ & 51.2$\%$  \\
        SQR      & IoU$>$0.75 & 21.1$\%$ & 47.0$\%$  \\
        \bottomrule
    \end{tabular}
    \caption{Baseline vs. SQR on true-positive fading rate and false-positive exacerbation rate. }
    \label{tab:ablation_FErates}
\end{table}

\subsection{Relation with Increased Number of Supervision}



Some concurrent studies reveal that query-based detectors suffer from training inefficiency due to the one-to-one matching, and propose to add extra parallel query groups and match them with ground-truths in a one-to-many manner, such as Group DETR \cite{Chen2022GroupDF} and H-DETR \cite{Jia2022DETRsWH}. We recognize them as \emph{increased number of supervision}, i.e., each ground-truth is matched to multiple queries, and thus the number of supervision ($\hat{q}$) is greatly increased. However, simply adding more query groups is a sub-optimal solution according to our motivation, since these extra supervisions are given uniformly among stages. So we present 6 designs as training strategies for investigation.

\begin{itemize}
    \item \textbf{Design I}: Following Group DETR, 3 groups of queries are initialized and independently undergo the whole process of the decoder. The pathway is 3 $\times$ $\mathcal{PT}^{1\text{-}2\text{-}3\text{-}4\text{-}5\text{-}6}$, and number of supervision is $ 3 $ groups $\times 6$ stages $= 18$.
    \item \textbf{Design II}: 4 groups of queries are initialized but the pathways are
    $\mathcal{PT}^{1\text{-}2\text{-}3\text{-}4\text{-}5\text{-}6}$, 
    $\mathcal{PT}^{1\text{-}2\text{-}3\text{-}4\text{-}5}$, 
    $\mathcal{PT}^{1\text{-}2\text{-}3\text{-}4}$, 
    $\mathcal{PT}^{1\text{-}2\text{-}3}$, so, with training emphasis on early stages, the number of supervision is $4+4+4+3+2+1=18$.
    \item \textbf{Design III}: Similar to Design II but not every initialized queries starts from stage 1. The pathways for the 4 groups are $\mathcal{PT}^{1\text{-}2\text{-}3\text{-}4\text{-}5\text{-}6}$,
    $\mathcal{PT}^{2\text{-}3\text{-}4\text{-}5\text{-}6}$, 
    $\mathcal{PT}^{3\text{-}4\text{-}5\text{-}6}$,
    $\mathcal{PT}^{4\text{-}5\text{-}6}$, so, with training emphasis on later stages, the number of supervision is $1+2+3+4+4+4=18$.
    \item \textbf{Design IV}: $SQR^{starting~stage = 2}$, the number of supervision is $1+1+2+3+5+8=20$.
    \item \textbf{Design V}: Same as Design I but with 6 groups. The number of supervision is $ 6 $ groups $\times 6$ stages $= 36$.
    \item \textbf{Design VI}: $SQR^{starting~stage = 1}$, the number of supervision is $1+2+3+5+8+13=32$.
\end{itemize}

\begin{table}[t!]
    \centering
    \begin{tabular}{l|c|c|c}
    \toprule[1pt]
        Design &  \#$Supv$ / stage & \#$Supv$ & AP \\ \midrule
        I~~~~(Group DETR) & 3,3,3,3,3,3 & 18 & 43.4 \\ 
        II & 4,4,4,3,2,1 & 18 & 43.0 \\ 
        III & 1,2,3,4,4,4 & 18 & 43.7 \\ 
        IV (SQR) & 1,1,2,3,5,8 & 20 & 44.2 \\ 
        \midrule
        V~~(Group DETR) & 6,6,6,6,6,6 & 36 & 43.6 \\ 
        VI (SQR)& 1,2,3,5,8,13& 32 & 44.4 \\ 
       \bottomrule
    \end{tabular}
    \caption{Results of the 6 designed training strategies on Adamixer to investigate the relation with number of supervision. The inference is untouched. \#$Supv$ denotes the number of supervision.}
    \label{tab:num_supervision}
\end{table}

Table.\ref{tab:num_supervision} summarizes the results. From the controlled experiments with Design I, II, III, we conclude training with emphasis on late stages is the best option. Design IV and VI are SQR with different starting stages.  We show that with similar or even fewer supervisions, SQR outperforms other designs by a large margin. These comparisons verify our motivation and prove that the enhancement from SQR is not simply from the increased number of supervisions.

\subsection{Training Emphasis via Re-weighted Loss}
Although \textit{the training emphasis} contributes to the success of SQR, the \textit{access to the early intermediate queries} is irreplaceable. We attempt another way to place training emphasis without query recollection, i.e. re-weighting the losses of six decoding stages following the Fibonacci sequence (1,2,3,5,8,13). The AP slightly degrades by (-0.6) compared to the baseline, indicating access to the intermediate queries is necessary for emphasized supervision.

\subsection{Relation with Stochastic Depth}
Stochastic Depth (SD) \cite{Huang2016DeepNW} is a training strategy for very deep network where it randomly removes a fraction of layers independently for each mini-batch, so the depth of pathway varies during training just like SQR. However, SD is of vital inefficiency, because an image has to first pass through a backbone and then be processed by a varying-depth pathway, but for SQR, an image that passes through the backbone can be processed by multiple varying-depth pathways at the same time. To verify this point, we apply SD on the vanilla decoder: During training, for each mini-batch we randomly remove decoding stages following a series of pre-defined probabilities. From stage 1$\sim$ 6, the removal probability $\mathbb{R}$ could be either a constant (0.1), increasing (0.0, 0.1,..., 0.5), or decreasing (0.5, 0.4,..., 0.0). During testing, SD requires the outputs of each decoding stage to be calibrated by the expected times the stage participates in training, so we multiply (1-$\mathbb{R}$) with the residual term in Equ.\ref{eq:1}. As a result, SD (40.7 mAP) is not comparable to SQR. 


\subsection{Training efficiency}

SQR leads to an extra computing burden compared to the baseline since the number of $\hat{q}$ grows. In Table.\ref{tab:startingstage}, SQR increases training time of Adamixer-R50 by 0.35$\sim$2.85 hours ($+0.07\%\sim 57\%$). In practice, the effects on different models/implementations/platforms vary. For instance, ResNet-50 is a light-weighted backbone, so the effects of applying SQR are relatively high, while with heavier backbones like ResNet-101 and Swin Transformer\cite{Liu2021SwinTH}, the training time is less impacted (+10$\%$). 

\begin{figure}[]
    \includegraphics[width=1\columnwidth]{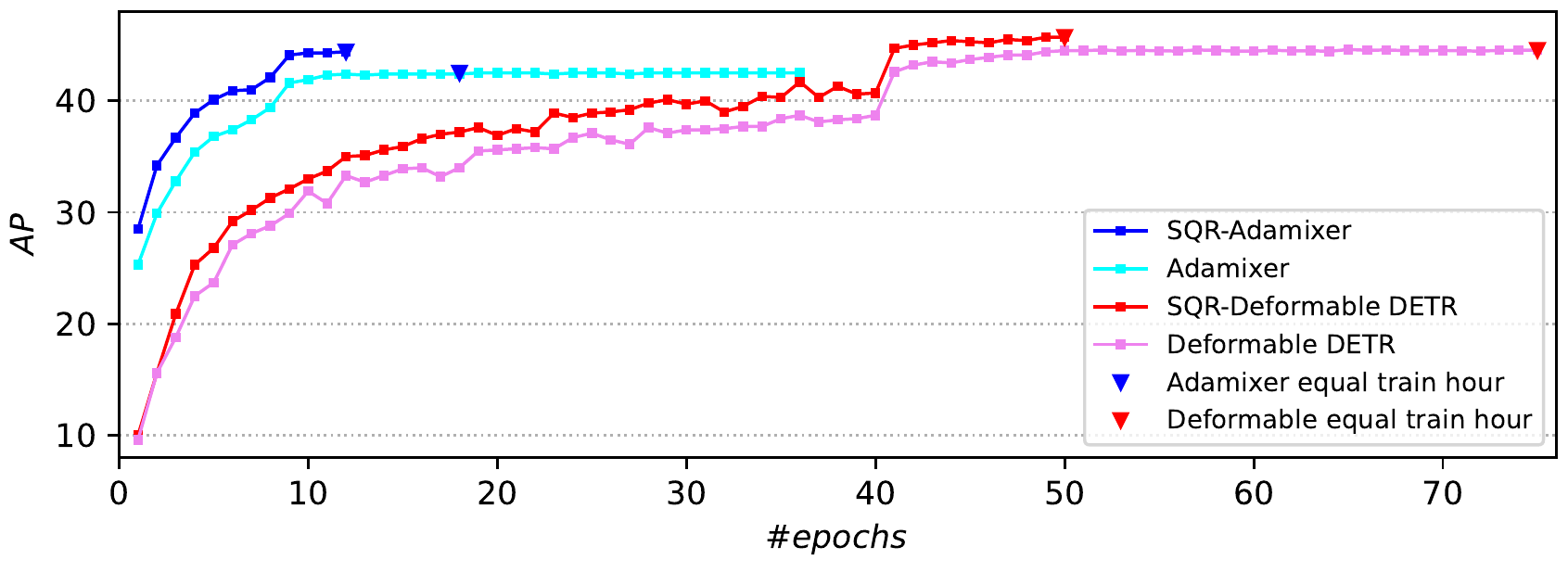}
    \caption{Equal training-time comparison: Baseline vs. SQR}
    \label{fig:training_time}
\end{figure}

For comparison under equal training time, we train the baseline Adamixer with extra epochs (+ 200$\%$ epochs) and baseline Deformable DETR with (+50$\%$ epochs). Both models' performances (see Fig.\ref{fig:training_time}) saturate early and there is no significant improvement over the extended training time, indicating that the benefit of SQR cannot be compensated by including more training iterations.

\subsection{Comparison with State-of-the-art}

\begin{table*}[t]
    \centering
    \begin{tabular}{@{}lccccccccc@{}}
    \toprule[1pt]
    \multicolumn{1}{l}{\multirow{2}{*}{Model}} &
    \multicolumn{1}{l}{\multirow{2}{*}{w/ SQR}} &
    \multicolumn{1}{c}{\multirow{2}{*}{\#query}}  & \multicolumn{1}{c}{\multirow{2}{*}{\#epochs}} & \multicolumn{6}{c}{COCO 2017 validation split} \\ \cmidrule(l){5-10}
     & &  &  & AP   & AP\textsubscript{50}    & AP\textsubscript{75}  & AP\textsubscript{S}   & AP\textsubscript{M}   & AP\textsubscript{L} \\ \midrule
    DETR-R50 \cite{Carion2020EndtoEndOD}      &          & 100   & 500   & 42.0  & 62.4  & 44.2  & 20.5  & 45.8  & 61.1  \\
    Conditional DETR-R50 \cite{Meng2021ConditionalDF} &  & 300   & 50    & 40.9  & 61.8  & 43.3  & 20.8  & 44.6  & 60.2  \\
    Conditional DETR-R101 \cite{Meng2021ConditionalDF} &  & 300   & 50    & 42.8  & 63.7  & 46.0  & 21.7  & 46.6  & 60.9  \\
    Anchor-DETR-R50 \cite{Wang2022AnchorDQ}   &          & 300   & 50    & 42.1  & 63.1  & 44.9  & 22.3  & 46.2  & 60.0  \\
    Anchor-DETR-R101 \cite{Wang2022AnchorDQ}  &          & 300   & 50    & 43.5  & 64.3  & 46.6  & 23.2  & 47.7  & 61.4  \\
    SAM-DETR-R50 \cite{Zhang2022AcceleratingDC}   &                      & 300   & 50    & 39.8  & 61.8  & 41.6  & 20.5  & 43.4  & 59.6  \\
    \textsuperscript{$*$}SMCA-DETR-R50 \cite{Gao2021FastCO} & & 300   & 50    & 43.7  & 63.6  & 47.2  & 24.2  & 47.0  & 60.4  \\
    \textsuperscript{$*$}SMCA-DETR-R50 \cite{Gao2021FastCO} & & 300   & 108   & 45.6  & 65.5  & 49.1  & 25.9  & 49.3  & 62.6  \\
    \textsuperscript{$*$}DN-DAB-DETR-R50 \cite{Li2022DNDETRAD} &                  & 300   & 50    & 44.1  & 64.4  & 46.7  & 22.9  & 48.0  & 63.4 \\ 
    \textsuperscript{$*$}DN-DAB-DETR-R101 \cite{Li2022DNDETRAD}   &               & 300   & 50    & 45.2  & 65.5  & 48.3  & 24.1  & 49.1  & 65.1 \\ \midrule 
    \textsuperscript{$*$}DAB-DETR-R50 \cite{Liu2022DABDETRDA}     &           & 300   & 50    & 42.2  & 63.1  & 44.7  & 21.5  & 45.7  & 60.3 \\
    \cellc \textsuperscript{$*$}SQR-DAB-DETR-R50 & \checkmark & \cellc 300 & \cellc 50 & \cellc \textbf{44.5} (+2.3)& \cellc 64.4 & \cellc 47.5 & \cellc 24.8 & \cellc 48.6 & \cellc 61.7 \\
    \textsuperscript{$*$}DAB-DETR-SwinB \cite{Liu2022DABDETRDA} & & 300   & 50    & 49.0  & 71.0  & 53.0  & 29.6  & 53.8  & 68.3 \\
    \cellc\textsuperscript{$*$}SQR-DAB-DETR-SwinB & \checkmark & \cellc300   & \cellc50    & \cellc \textbf{51.6} (+2.6) & \cellc 72.5 &\cellc 55.9 & \cellc 32.0 & \cellc 56.8 & \cellc 71.0 \\ \midrule
    \textsuperscript{$*$}Deformable DETR-R50 \cite{Zhu2021DeformableDD}  &    & 300   & 12    & 37.2  & 55.2  & 40.4  & 20.6  & 40.2  & 50.2 \\
    \cellc \textsuperscript{$*$}SQR-Deformable DETR-R50 & \checkmark & \cellc 300 & \cellc 12 & \cellc \textbf{39.9} (+2.7) & \cellc 58.4 & \cellc 43.7 & \cellc 23.8 & \cellc 43.2 & \cellc 53.3 \\
    \textsuperscript{$*$}Deformable DETR-R50 \cite{Zhu2021DeformableDD}   &   & 300   & 50    & 44.5  & 63.2  & 48.9  & 28.0  & 47.8  & 58.8 \\
    \cellc \textsuperscript{$*$}SQR-Deformable DETR-R50   &    \checkmark   & \cellc 300 & \cellc 50 & \cellc \textbf{45.9} (+1.4) & \cellc 64.7 & \cellc 50.2 & \cellc 27.7 & \cellc 49.2 & \cellc 60.5 \\ \midrule
    Adamixer-R50 \cite{Gao2022AdaMixerAF}        &       & 100   & 12    & 42.5  & 61.5  & 45.6  & 24.6  & 45.1  & 59.2 \\
    \cellc SQR-Adamixer-R50               &  \checkmark & \cellc 100 & \cellc 12 & \cellc \textbf{44.4} (+1.9) & \cellc 63.2 & \cellc 47.8 & \cellc 25.7 & \cellc 47.4 & \cellc 60.2 \\
    \textsuperscript{$\dagger$}Adamixer-R50 \cite{Gao2022AdaMixerAF}  &  & 100   & 12    & 42.5  & 61.5  & 45.8  & 24.4  & 45.2  & 58.7 \\ 
    \cellc \textsuperscript{$\dagger$}SQR-Adamixer-R50      &  \checkmark   & \cellc 100 & \cellc 12 & \cellc \textbf{45.3} (+2.8) & \cellc 63.8 & \cellc 49.0 & \cellc 26.8 & \cellc 48.1 & \cellc 62.2 \\
    \textsuperscript{$*\dagger$}Adamixer-R50    &   &  100 & 36 & 45.1 & 63.9 & 48.9 & 28.3 & 47.8 & 60.6 \\
    \cellc \textsuperscript{$*\dagger$}SQR-Adamixer-R50    & \checkmark  & \cellc 100 & \cellc 36 & \cellc \textbf{46.7} (+1.6) & \cellc 65.2 & \cellc 50.3 & \cellc 29.4 & \cellc 49.6 & \cellc 62.1 \\ 
    \textsuperscript{$*\dagger$}Adamixer-R50   &    & 300 & 36 & 46.6 & 65.5 & 50.6 & 29.3 & 49.3 & 62.3 \\
    \cellc \textsuperscript{$*\dagger$}SQR-Adamixer-R50  & \checkmark & \cellc 300 & \cellc 36 & \cellc \textbf{48.9} (+2.3) & \cellc 67.5 & \cellc 53.2 & \cellc 32.0 & \cellc 51.8 & \cellc 63.7 \\
    \textsuperscript{$*\dagger$}Adamixer-R101  \cite{Gao2022AdaMixerAF} & &100   & 36 & 45.7 & 64.7 & 49.6 & 27.8 & 49.1 & 61.2 \\ 
    \cellc \textsuperscript{$*\dagger$}SQR-Adamixer-R101 & \checkmark & \cellc 100   & \cellc 36 & \cellc \textbf{47.3} (+1.6) & \cellc 66.0 & \cellc 51.3 & \cellc 30.1 & \cellc 50.7 & \cellc 62.2 \\
    \textsuperscript{$*\dagger$}Adamixer-R101 \cite{Gao2022AdaMixerAF} & & 300   & 36 & 47.6 & 66.7 & 51.8 & 29.5 & 50.5 & 63.3 \\  
    \cellc \textsuperscript{$*\dagger$}SQR-Adamixer-R101 & \checkmark& \cellc 300   & \cellc 36 & \cellc \textbf{49.8} (+2.2) & \cellc 68.8 & \cellc 54.0 & \cellc 32.0 & \cellc 53.4 & \cellc 65.1 \\ 
    \bottomrule[1pt]
    \end{tabular}
    \caption{Comparison results with various query-based detectors on COCO 2017 val. \#query: the number of queries used during inference. * indicates models trained with multi-scale augmentation, $\dagger$ marks models with 7 decoder stages.}
    \label{tab:sota}
\end{table*}

We conduct experiments on recent query-based detectors with different training settings and various backbones, with and without SQR. We primarily follow the original training setting of our baselines, where training schedules consists of standard 12, 36, and 50 epochs; the number of queries is chosen between 100 and 300; multi-scale training is applied to 36e and 50e schedules as the shorter side of images range 480 $\sim$ 800. The training is conducted on 8x Nvidia A100.

The result is summarized in Fig.\ref{fig:speedvsap} and Table \ref{tab:sota}. SQR consistently brings AP improvements to Adamixer, DAB-DETR and Deformable-DETR at the same inference speed. Concretely, on DAB-DETR, SQR brings +2.3 and +2.6 AP with R50 and SwinB, respectively; on Deformable DETR, SQR boosts it by 2.7 AP under 12e and by 1.4 AP under 50e; on Adamixer with R50, SQR achieves +1.9 AP under the basic setting (100 queries, 12e). With an additional stage, the gap between w/ and w/o SQR is enlarged +2.8 AP. SQR consistently improves these models by 1.4$\sim$2.8 AP. 




\subsection{DQR with Recurrence for Reducing Model Size}
Herein, we explore an interesting direction enabled by Dense Query Recollection, i.e., using DQR to reduce model size. Existing methods typically have more than 6 decoding stages in decoder. Can we directly train a detector where all decoding stages share parameters?  We implement this concept on vanilla Adamixer and find that the model is not able to converge. But we find DQR has the capability to achieve the goal.

As we know, a strong decoding stage at the end, i.e. the final stage, is obtained after training with DQR. This stage has seen every possibly intermediate queries that ever exist along the decoding process. A natural attempt is to replace all stages' parameters with the final stage's parameter during inference, forming a pathway as $\mathcal{PT}^{6\text{-}6\text{-}6\text{-}6\text{-}6\text{-}6}$. However, this results in a 0 AP result! The reason is that the output of stage $6$ shifts from its input, so stage $6$ cannot recognize its own output, thus, it applies random refinement (negative effect) on it.

To address the problem, during training, \textbf{we recollect the output of stage $6$, and feed back to itself as its input}. In such way, stage 6 gets chance to learn refining its output. Then, we recurrently use stage 6 only for inference. We name this method as DQRR (Dense Query Recollection and Recurrence). The result is shown in Table \ref{tab:DQRR}.

\begin{table}[t!]
    \centering
    \begin{tabular}{l|c|c|c}
    \toprule[1pt]
        \# stage  &  AP & AP50 & AP75 \\
        \midrule
        1       & 0.125  & 0.290  &  0.092  \\
        2       & 0.329  & 0.514  &  0.346  \\
        3       & 0.400  & 0.583  &  0.427  \\
        4       & 0.422  & 0.606  &  0.453  \\
        5       & 0.428  &  0.612 &  0.459  \\
        6       & 0.428  &  0.613 &  0.459 \\
       \bottomrule
    \end{tabular}
    \caption{Dense query recollection and recurrence.}
    \label{tab:DQRR}
\end{table}

With DQRR, all decoding stages share the same parameters, so the model size is reduced by 70\% (1.6GB to 513 MB). And it only needs 5 stages to achieves better performance than previous (42.8AP vs 42.5 AP).

\section{Conclusion}
\label{sec:conclusion}
In this work, we investigate the phenomenon where the optimal detections of query-based object detectors are not always from the last decoding stage, but can sometimes come from an intermediate decoding stage. We first recognize two limitations causing the issue, i.e. lack of training emphasis and cascading errors from query sequence. The problem is addressed by Selective Query Recollection (SQR) as a simple and effective training strategy.
Across various training settings, SQR boosts Adamixer, DAB-DETR, and Deformable-DETR by a large margin.

{\small
\bibliographystyle{ieee_fullname}
\bibliography{egbib}
}

\newcommand{\beginsupplement}{%
        \setcounter{table}{0}
        \renewcommand{\thetable}{S\arabic{table}}%
        \setcounter{figure}{0}
        \renewcommand{\thefigure}{S\arabic{figure}}%
     }

\appendix
\beginsupplement
\section{Implementation Detail}
\subsection{Query with Priors}
Recent query-based object detectors associate priors with queries.  These priors have multiple forms \cite{Gao2022AdaMixerAF,Liu2022DABDETRDA, Wang2022AnchorDQ, Meng2021ConditionalDF}, but generally, they are designed to involve spatial and scale priors which helps models converge faster. When implementing these methods, a query is usually regarded as two parts: one is the embedding that focuses on interacting with feature map and producing high-level object information, called \emph{content}, another is similar to the concept \emph{anchor}\cite{Ren2015FasterRT} which serves as a \emph{reference} point for locating/scaling the object, and narrows down the range of feature-interaction (e.g., cross-attention). 

For SQR, we recollect the \emph{query} at each stage. That means both the content and the corresponding reference are recollected in the our operation.

\section{Additional Experiments And Analysis}

\subsection{SQR with DN-DETR}
\begin{table}[h]
    \centering
    \begin{tabular}{l|c|c|c|c}
    \toprule[1pt]
        Model &  Epoch &  AP & AP50 & AP75 \\ 
        \midrule
        DN-DETR \cite{Li2022DNDETRAD}        & 12e & 38.5 & 58.8 & 40.6 \\ 
        SQR-DN-DETR    & 12e & \textbf{40.4} & 61.1 & 42.7 \\
        \midrule
        DN-DETR \cite{Li2022DNDETRAD}        & 50e & 44.1 & 64.4 & 46.7 \\ 
        SQR-DN-DETR    & 50e & \textbf{45.2} & 65.7 & 48.3 \\
       \bottomrule
    \end{tabular}
    \caption{SQR with DN-DETR}
    \label{tab:SQR-DN-DETR}
\end{table}

DN-DETR \cite{Li2022DNDETRAD} is a training strategy that is based on DAB-DETR. We show that SQR is compatible with DN. Table \ref{tab:SQR-DN-DETR} presents the results of DN-DETR w/ and w/o SQR. SQR enhances DN-DETR by $+1.9$ with 12 epochs schedule, and $+1.1$ with 50 epochs. We see that the benefit of SQR is less with extra long training epochs than with standard 1x schedule, although the improvement is still significant. Similar observations are obtained from Table {\color{red} 8} as Deformable DETR get $+2.7$ AP by SQR with 12 epochs while get $+1.4$ AP with 50 epochs. We elaborate this point in the following section.

\subsection{SQR with 12e And 50e}

Observation on Table \ref{tab:SQR-DN-DETR} and Table {\color{red} 8} indicate that the benefit of SQR becomes less with extra long training epochs than with standard 1x schedule. However, the benefit cannot be replaced by extended training epochs, as already analyzed in Fig.{\color{red} 5}. DN-DETR is under-fitted at the 12th epoch because of its limited convergence speed and the multi-scale training setting, in this case, the mechanism of query recollection produces more supervision and speeds up the convergence of later stages. Under 50 epochs, as later stages get more supervisions and the model converges, the benefit of accelerated convergence becomes less, on the other hand, the benefit from the training emphasis and the mitigated cascading errors is less affected by the long schedule, which still brings strong improvement.

\section{Notation}

The notation used in this paper is summarized in Table \ref{tab:notation}
\begin{table}[t]
    \centering
    \caption{Notation in this paper by the order of appearance.}
    \begin{tabular}{@{}ll@{}}
    \toprule[1pt]
    Notation & Definition \\ 
    \midrule
    QR      &   Query Recollection   \\
    SQR    &  Selective Query Recollection   \\
    $q_i^0$    &   The $i_{th}$ initial query      \\
    n    &   Total number of initial query  \\
    N    & The set \{1,2,3, ..., n\}\\
    $q_i^1$ & The output of the first stage refining $q_i^0$,\\
           & also known as the $i_{th}$ query at stage 1 \\
    s & The index of stage\\
    $D^s$ & The $s_{th}$ decoding stage\\
    $q_i^s$ & The $i_{th}$ query at stage s \\
    $q^s$ & The set of queries $q^s=\{q_i^s|i\in N\}$\\
    $x$ & Image features\\
    $(\mathcal{A}\circ\mathcal{F})$ & Self- and cross-attention and feed forward network\\
    $P_i^s$ & Prediction of $q_i^s$\\
    G & A ground-truth \\
    IoU & Intersection over Union\\
    TP & True-positive\\
    FP & False-positive\\
    DQR & Dense Query Recollection\\
    $q$ & A set of queries $\{q_{i}|i\in \{1,2,...,n\}\}$, as a basic unit\\
    $\hat{q}$ & $q$ that requires supervision during training\\
    $\mathcal{PT}$ & Pathway\\
    $C$ & A collection of $q$\\
    $\#Supv$ & Number of supervision\\
    $\mathbb{R}$ & Removal Probability\\
    DQRR & Dense query recollection and recurrence  \\
    \bottomrule[1pt]
    \end{tabular}
    \label{tab:notation}
\end{table}

\end{document}